\def\year{2016}\relax
\documentclass[letterpaper]{article}
\usepackage{aaai16_accepted}
\usepackage{times}
\usepackage{helvet}
\usepackage{courier}
\usepackage{epsfig}
\usepackage{graphicx}
\usepackage{amsmath}
\usepackage{amssymb}
\usepackage{algorithmic}
\usepackage{algorithm}
\usepackage{memhfixc}
\usepackage{amstext}
\usepackage{fixltx2e}
\usepackage{paralist}
\usepackage{placeins}
\usepackage{fancyhdr}
\usepackage{booktabs}

\DeclareMathOperator*{\argmin}{argmin}
\DeclareMathOperator*{\argmax}{argmax}

\begin{document}
%
\title{Labeling the Features Not the Samples: \\
        Efficient Video Classification with Minimal Supervision}

\author{Marius Leordeanu$^{1,2}$  \;\;   Alexandra Radu$^{1,2}$  \;\;   Shumeet Baluja$^{3}$  \;\;  Rahul Sukthankar$^{3}$\\
        $^{1}$Institute of Mathematics of the Romanian Academy, Bucharest, Romania \\
        $^{2}$University Politehnica of Bucharest, Bucharest, Romania \\
        $^3$Google Research, Mountain View, CA, USA \\
}
\maketitle

\begin{abstract}
\begin{quote}
Feature selection is essential for effective visual recognition. We propose an efficient joint classifier learning and feature selection method
that discovers sparse, compact representations of input features from a vast sea of candidates, with an almost
unsupervised formulation. Our method
requires only the following knowledge, which we call the \emph{feature sign}---whether or not a particular feature has on average
stronger values over positive samples than over negatives.
We show how this can be estimated using as few as a single labeled training sample per class. Then, using
these feature signs, we extend an initial supervised learning problem into an (almost)
unsupervised clustering formulation that can
incorporate new data without requiring ground truth labels.
Our method works both as a feature selection mechanism and as a fully competitive classifier.
It has important properties, low computational cost and excellent accuracy, especially in difficult cases
of very limited training data.
We experiment on large-scale recognition in video and show superior speed and performance to established feature selection approaches such as AdaBoost, Lasso, greedy forward-backward selection, and powerful classifiers such as SVM.
\end{quote}
\end{abstract}

\section{Introduction}

\begin{figure}
\begin{center}
\includegraphics[scale = 0.27, angle = 0, viewport = 0 0 900 740, clip]{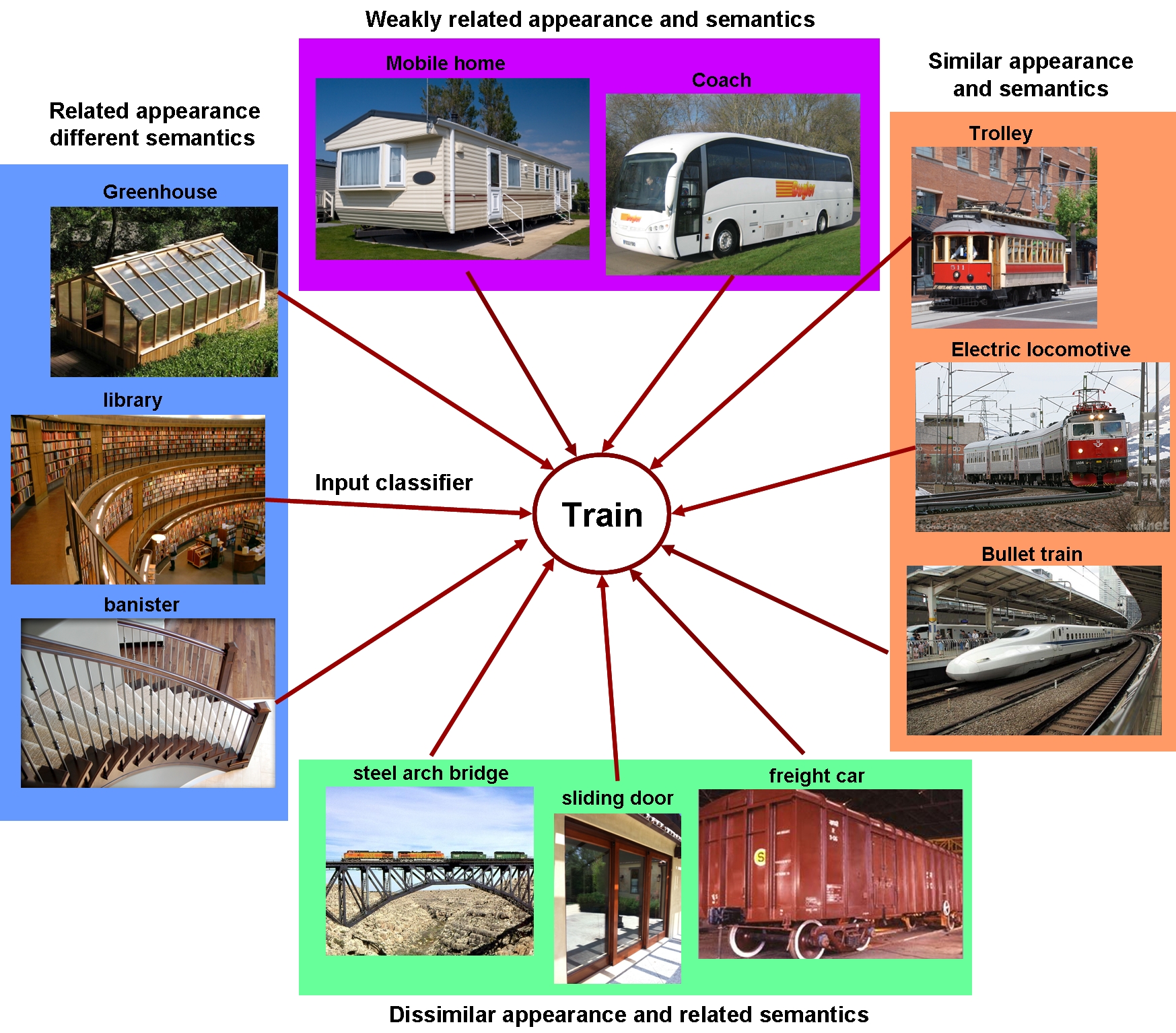}
\caption{What classes can \emph{trigger} the idea of
a ``train''? Many classes have similar appearance but are semantically less related (blue box);
others are semantically closer but visually less similar (green box). There is a continuum that relates appearance, context and semantics. Can we find a
group of classifiers, which
are together robust to outliers, over-fitting and missing features?
Here, we show classifiers that are consistently
selected by our method from limited training data as
giving valuable input to the class ``train''.
}
\label{fig:teaser_train}
\end{center}
\end{figure}

Classifier ensembles and feature selection have
proved enormously useful over decades of computer vision
and machine learning
research
~\cite{vasconcelos2003feature,dietterich2000ensemble,hansen1990neural,breiman1996bagging,freund1995decision,kwok2013multiple,criminisi2012decision}.
Every year, new visual features and classifiers
are proposed or automatically learned.
As the vast pool of features continues to grow, efficient feature selection
mechanisms must be devised since classes are often
\emph{triggered} by only a few key input features
(Fig.~\ref{fig:teaser_train}).
As feature selection is NP-hard \cite{guyon2003introduction,ng1998feature},
previous work focused on greedy methods, such as sequential search
\cite{pudil1994floating} and boosting \cite{freund1995decision},
relaxed formulations with $l^1$- or $l^2$-norm regularization, such as ridge regression \cite{vogel2002computational} and
the Lasso \cite{tibshirani1996regression,zhao2006model}, or heuristic genetic algorithms \cite{siedlecki1989note}.

We approach feature selection from the task of discriminant linear
classification \cite{DuHa73} with novel constraints on the solution and the features.
We put an upper bound on the solution weights and require it to
be an affine combination of soft-categorical features,
which should have on average stronger outputs on the positive class vs. the negative.
We term these \emph{signed features}.
We present both a supervised and an almost unsupervised
approach.
Our supervised method is a convex constrained
minimization problem, which we extend to the case of
almost unsupervised learning, with a concave minimization formulation,
in which the only bits of
supervised information required are the feature signs.
Both formulations
have important sparsity and optimality properties as well as strong generalization
capabilities in practice.
The proposed schemes also serve as feature
selection mechanisms,
such that the majority of features with zero weights can be safely ignored
while the remaining ones form a powerful classifier ensemble.
Consider Fig.~\ref{fig:teaser_train}: here we use image-level CNN classifiers \cite{jia2014caffe}, pre-trained on ImageNet, to recognize trains in video frames from the YouTube-Objects dataset \cite{prest2012learning}.
Our method rapidly finds relevant
features in a large pool.

Our main contributions are:
1) An efficient method for joint linear classifier learning and
feature selection.
We show that, both in theory and practice, our solutions are sparse.
The number of features selected can be set to $k$ and the non-zero weights are equal to $1/k$. The simple solution enables good generalization and learning
in an almost unsupervised setting, with minimal supervision.
This is very different from classical regularized approaches such as the Lasso.
2) Our formulation requires minimal supervision: namely only the \emph{signs} of features with respect to the target class.
These signs can be estimated from a small set of labeled samples, and once determined,
our method can handle large quantities of unlabeled data with excellent accuracy and generalization in practice.
Our method is also robust to large errors in feature sign estimation.
3) Our method demonstrates superior performance in terms of learning time and accuracy when compared to established approaches such as AdaBoost, Lasso, Elastic Net and SVM, especially in the case of limited supervision.

\section{Problem Formulation}
\label{sec:formulation}

We address the case of binary classification, and apply the one vs.\ all
strategy to the multi-class scenario.
Consider a set of $N$ samples, with each $i$-th sample expressed
as a column vector $\mathbf{f}_i$ of $n$ features with values in $[0,1]$;
such features could themselves be outputs of classifiers.
We want to find
vector $\mathbf{w}$, with elements in $[0,1/k]$ and unit $l^1$-norm,
such that $\mathbf{w}^T\mathbf{f}_i \approx \mu_P$ when the $i$-th sample is
from the positive class and $\mathbf{w}^T\mathbf{f}_i \approx \mu_N$ otherwise, with $0 \leq \mu_N < \mu_P \leq 1$.
For a labeled training sample $i$, we
fix the ground truth target $t_i = \mu_P = 1$
if positive and $t_i = \mu_N = 0$ otherwise.
Our novel constraints on $\mathbf{w}$
limit the impact of each individual feature $f_j$, encouraging the selection of features that are powerful in combination, with no single one strongly dominating. This produces solutions with good generalization power.
In a later section we
show that $k$ is equal to the number of selected features, all with
weights $=1/k$. The solution we look for is
a weighted feature average
with an ensemble response that is stronger on positives than on negatives.
For that, we want any feature $f_j$ to have expected value $E_P(f_j)$ over positive samples
greater than its expected value $E_N(f_j)$ over negatives.
We estimate its sign $\mathit{sign}(f_j) = E_P(f_j) - E_N(f_j)$
from labeled samples and if it is negative we simply \emph{flip} the feature: $f_j \gets 1 - f_j$.
Expected values are estimated as the empirical average feature responses the
the labeled training data available.

\paragraph{Supervised Learning:}
We begin with the supervised learning task, which
we formulate as a least-squares constrained minimization problem.
Given the $N \times n$ feature matrix $\mathbf{F}$ with
$\mathbf{f}^\top_i$ on its $i$-th row and the ground truth vector $\mathbf{t}$, we look for
$\mathbf{w^*}$ that minimizes
$\|\mathbf{Fw} - \mathbf{t}\|^2=\mathbf{w^\top (F^\top F) w} - 2\mathbf{(F^\top t)}^\top \mathbf{w} + \mathbf{t}^\top \mathbf{t}$, and obeys the required constraints. We drop
the last constant term $\mathbf{t}^\top \mathbf{t}$
and obtain the following convex minimization problem:
\begin{eqnarray}
\label{eq:learning}
\mathbf{w^*} & = & \argmin_w J(\mathbf{w}) \\
    & = & \argmin_w \mathbf{w^\top (F^\top F) w} - 2\mathbf{(F^\top t)}^\top \mathbf{w} \nonumber \\
    & & s.t. \sum_j w_j =1 \;, \; w_j \in [0,1/k]. \nonumber
\end{eqnarray}
\noindent
The least squares formulation is related to Lasso, Elastic Net and other regularized approaches, with the
distinction that in our case
individual elements of $\mathbf{w}$ are restricted to $[0,1/k]$. This
leads to important properties regarding sparsity
and directly impacts generalization power,
as presented later.

\paragraph{Labeling the features not the samples:}

Consider a pool of signed features correctly flipped according to their signs, which could be known a priori, or estimated from a small set of
labeled data. We make the simplifying assumption that the signed features'
expected values (that is, the means of the feature responses distributions), for positive and negative samples, respectively,
are close to the ground truth target values
$(\mu_P, \mu_N)$. Note that having expected values close to the ground truth
does not say anything about the distribution variance, as individual responses
could sometimes be wrong.
For a given sample $i$, and any
$\mathbf{w}$ obeying the constraints, the expected value of the weighted average $\mathbf{w}^\top\mathbf{f}_i$ is also close to the ground truth target $t_i$:
$E(\mathbf{w}^\top\mathbf{f}_i)=\sum_j w_j E(\mathbf{f}_i(j))
\approx (\sum_j w_j)t_i = t_i$.
Then, for all samples we have the expectation
$E(\mathbf{F}\mathbf{w}) \approx \mathbf{t}$, such that any feasible
solution will produce, on average, approximately correct answers.
Thus, we can regard the supervised learning scheme as attempting to reduce the variance of the feature ensemble output, as their expected value
is close to the ground truth target. If we approximate
$E(\mathbf{F}\mathbf{w}) \approx \mathbf{t}$ into the
objective $J(\mathbf{w})$, we get a new ground-truth-free objective $J_u(\mathbf{w})$ with the following
learning scheme, which is unsupervised once the feature signs have been estimated.
Here $\mathbf{M = F^\top F}$:
\begin{eqnarray}
\label{eq:semisup_learning}
\mathbf{w^*} & = & \argmin_w J_u(\mathbf{w} )\\
    & = & \argmin_w \mathbf{w^\top (F^\top F) w} - 2\mathbf{(F^\top (\mathbf{F}\mathbf{w}))}^\top \mathbf{w} \nonumber \\
    & = & \argmin_w (-\mathbf{w^\top (F^\top F w})) = \argmax_w \mathbf{w^\top M w} \nonumber \\
    & & \text{s.t.} \sum_j w_j =1 \;, \; w_j \in [0,1/k]. \nonumber
\end{eqnarray}
Interestingly, while the supervised case is a convex minimization problem,
the semi-supervised learning scheme
is a concave minimization problem, which is NP-hard.
This is due to the change in sign of the matrix $\mathbf{M}$.
Since $\mathbf{M}$ in the almost unsupervised
case could be
created from larger quantities of unlabeled data, $J_u(\mathbf{w})$ could
in fact be less noisy than $J(\mathbf{w})$ and produce significantly
better local optimal solutions --- a fact confirmed by experiments.
Note the difference between our formulation and other, much more costly semi-supervised
or transductive learning approaches based on
label propagation with quadratic criterion \cite{bengio2006label}
(where the quadratic term is very large, being
computed from pairs of data samples, not features) or on
transductive support vector machines \cite{joachims1999transductive}.
There are also methods for unsupervised feature selection,
such as the regularization scheme of \cite{yang2011l2}, but they do
not simultaneously learn a
discriminative classifier, as it is the case here.

\paragraph{Intuition:}

Let us consider two terms involved in our objectives,
the quadratic term: $\mathbf{w^\top M w}=\mathbf{w^\top (F^\top F) w}$
and the linear term: $\mathbf{(F^\top t)}^\top \mathbf{w}$.
Assuming that feature outputs have similar expected values,
then minimizing the linear term in the supervised case
will give more weight to features that are strongly correlated with the ground truth and are good
for classification, even independently.
Things become more interesting when looking at the role played by the quadratic term in the two cases of learning.
The positive definite
matrix $\mathbf{F^\top F}$ contains the dot-products between pairs of feature responses over the samples.
In the supervised case, minimizing $\mathbf{w^\top (F^\top F) w}$
should find \emph{groups of features}
that are as uncorrelated as possible. Thus, they should be
individually relevant due to the linear term,
but not redundant with respect to each other due to the quadratic term.
They should be \emph{conditionally independent} given the class, an observation that is consistent with earlier
research (e.g., \cite{dietterich2000ensemble,rolls2010noisy}).
In the almost unsupervised case, the task seems reversed: maximize the same
quadratic term $\mathbf{w^\top M w}$, with no linear term involved. We could interpret this as
transforming the learning problem into a special case of clustering with pairwise constraints,
related to methods such as spectral clustering with
$l^2$-norm constraints \cite{key:sarkar} and robust hypergraph clustering with
$l^1$-norm constraints \cite{key:bulo_nips09,key:latecki_nips10}.
The problem is addressed by finding the group of features with strongest
intra-cluster score --- the largest amount of covariance.
In the absence of
ground truth labels, if we assume that features in the pool are,
in general, correctly signed
and not redundant,
then the maximum covariance is attained by those
whose collective average varies the most as the hidden class labels also vary.



\section{Algorithm}
\label{sec:algorithms}

\begin{figure}[t!]
\begin{center}
\includegraphics[scale = 0.35, angle = 0, viewport = 0 0 550 520, clip]{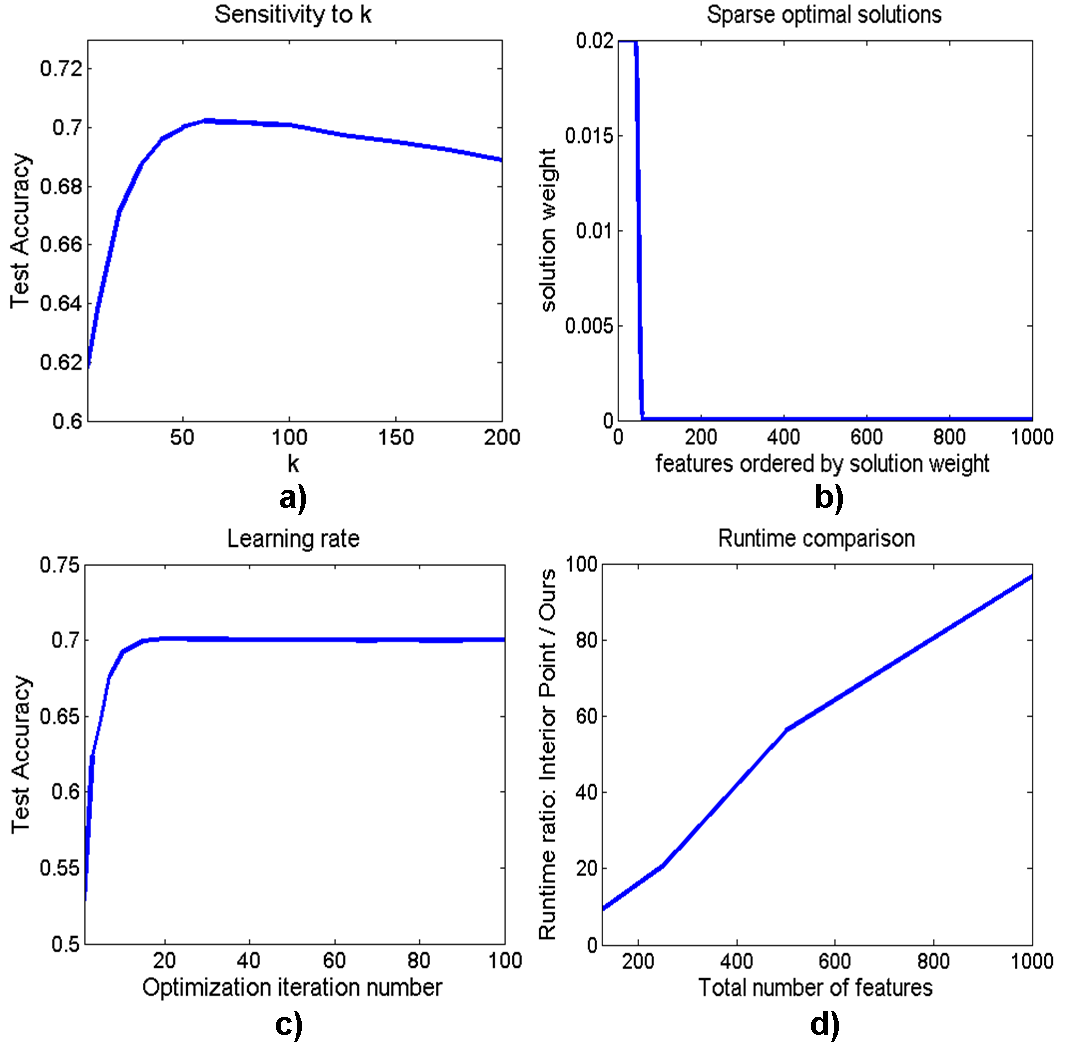}
\caption{Optimization and sensitivity analysis:
a) Sensitivity to $k$.
Performance improves as features are added, is stable
around the peak $k=60$ and falls for $k>100$ as useful features
are exhausted.
b) Features ordered by weight for $k=50$ confirming that our method selects
equal weights up to the chosen $k$.
c) Our method almost converges in $10$--$20$ iterations.
d) Runtime of interior point method divided by ours,
both in Matlab and with $100$ max iterations. All results are averages
over $100$ random runs.
}
\label{fig:optimization}
\end{center}
\end{figure}

We first need
to estimate the \emph{sign} for each feature,
using its average response over positives and negatives,
respectively. Then we can set up the optimization
problems to find $\mathbf{w}$. In Algorithm
\ref{alg:semisup_learning}, we present the almost unsupervised method,
with the supervised variant being constructed by modifying the objective
appropriately.
There are many possible fast methods for approximate optimization.
Here we adapted the integer projected
fixed point (IPFP) approach \cite{leordeanu2012efficient,key:leordeanu_IPFP},
which is efficient in practice (Fig.~\ref{fig:optimization}c)
and is applicable to both supervised and semi-supervised cases.
The method converges to a stationary point --- the
global optimum in the supervised case.
At each iteration IPFP approximates
the original objective
with a linear, first-order Taylor approximation that can be optimized immediately in the feasible domain.
That step is followed by a \emph{line search} with rapid closed-form solution, and the process is repeated
until convergence.
In practice, $10$--$20$
iterations bring us close to the stationary point;
nonetheless, for thoroughness, we use $100$ iterations
in all tests. See, for example,
comparisons to Matlab's \emph{quadprog} run-time
for the convex supervised learning case in Fig.~\ref{fig:optimization}
and to other learning methods in Fig.~\ref{fig:test_and_time}.
Note that once the linear and quadratic terms are set up, the learning
problems are independent of the number of samples and only dependent
on the number $n$ of features considered, since $\mathbf{M}$ is $n \times n$
and $\mathbf{F^\top t}$ is $n \times 1$.


\begin{algorithm}
\caption{Learning with minimal supervision.}
\label{alg:semisup_learning}
\begin{algorithmic}
\STATE Learn feature signs from a small set of labeled samples.
\STATE Create $\mathbf{F}$ with flipped features from unlabeled data.
\STATE Set $\mathbf{M} \gets \mathbf{F^\top F}$. \\
\STATE Find $\mathbf{w^*} = \argmax_w \mathbf{w^\top M w}$ \\
       $\; \; \; \; \; \; $ s.t. $\sum_j w_j =1 \;, \; w_j \in [0,1/k]$.
\RETURN $\mathbf{w^*}$
\end{algorithmic}
\end{algorithm}


\paragraph{Theoretical Analysis:}
First we show that the solutions are sparse
with equal non-zero weights (P1), also
observed in practice (Fig.~\ref{fig:optimization}b).
This property makes our classifier learning also
an excellent feature selection mechanism.
Next, we show that simple equal weight solutions are likely to minimize the output variance over samples of a given class
(P2) and minimize the error rate. This explains the
good generalization power.
Then we show how the error rate is expected
to go towards zero when the number of considered
non-redundant features increases (P3),
which explains why a large diverse pool of features is beneficial.
Let $J(\mathbf{w})$ be the objective for either the
supervised or semi-supervised case:

\begin{figure}
\begin{center}
\includegraphics[scale = 0.27, angle = 0, viewport = 0 0 700 280, clip]{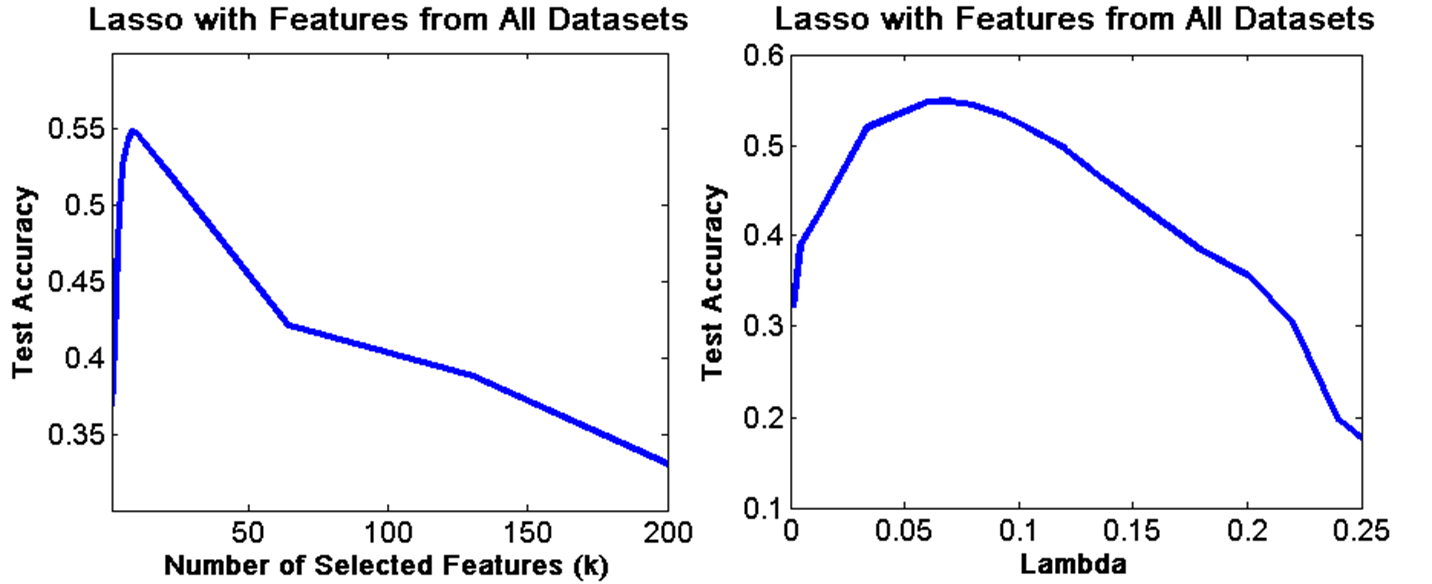}
\caption{Sensitivity analysis for Lasso:
Left: sensitivity to number of features with non-zero weights
in the solution.
Note the higher sensitivity when compared to ours.
Lasso's best performance is achieved
for fewer features, but the accuracy is worse than in our case.
Right: sensitivity to lambda $\lambda$, which controls the
L1-regularization penalty.
}
\label{fig:Lasso_sensitivity}
\end{center}
\end{figure}

\noindent \textbf{Proposition 1:}
Let $\mathbf{d(w)}$ be the gradient
of $J(\mathbf{w})$. The partial derivatives $d(\mathbf{w})_i$ corresponding
to those elements $w^*_i$ of the stationary points
with non-sparse, real values in $(0,1/k)$ must be
equal to each other.

\noindent \textbf{Proof:}
The stationary points for the Lagrangian
satisfy the Karush-Kuhn-Tucker (KKT) necessary optimality conditions.
The Lagrangian is $L(\mathbf{w}, \lambda, \mu , \beta) = J(\mathbf{w}) - \lambda (\sum w_i - 1) + \sum \mu_i w_i + \sum \beta_i (1/k - w_i)$.
From the KKT conditions at a point $\mathbf{w^*}$ we have:
\[
\begin{array}{l}
\mathbf{d(w^*)} - \lambda + \mu_i - \beta_i = 0,\\
\sum_{i=1}^{n} \mu_i w^*_i = 0,\\
\sum_{i=1}^{n} \beta_i (1/k - w^*_i) = 0.\\
\end{array}
\]
Here $\mathbf{w^*}$ and the Lagrange multipliers have non-negative elements,
so if $w_i > 0 \Rightarrow \mu_i = 0$
and $w_i < 1/k \Rightarrow \beta_i = 0$. Then there must exist
a constant $\lambda$ such that:
\[d(\mathbf{w^*})_i =  \left\{
\begin{array}{ll}
 \leq \lambda, & w^*_i = 0, \\
 = \lambda,    & w^*_i \in (0, 1/k), \\
\geq \lambda, & w^*_i = 1/k. \\
\end{array}\right.
\]
This implies that all $w^*_i$ that are different from $0$ or $1/k$
correspond to partial derivatives $d(\mathbf{w})_i$ that are equal
to some constant $\lambda$, therefore
those $d(\mathbf{w})_i$ must be equal to each other,
which concludes our proof.

From Proposition $1$ it follows that in the general case, when the
partial derivatives of the objective error function
at the Lagrangian stationary point are unique, the elements of
the solution $\mathbf{w^*}$ are either $0$ or $1/k$.
Since $\sum_j w^*_j = 1$ it follows that the number of nonzero weights
is exactly $k$, in the general case.
Thus, our solution is not just a simple linear
separator (hyperplane), but also a sparse representation and a feature selection
procedure that effectively averages the selected $k$ (or close to $k$) features.
The method is robust to the choice of $k$ (Fig.~\ref{fig:optimization}.a)
and seems to be less sensitive to the number of features selected
than the Lasso (see Fig.~\ref{fig:Lasso_sensitivity}).
In terms of memory cost,
compared to the solution
with real weights for all features, whose storage requires
$32n$ bits in floating point representation, our averaging of $k$ selected
features needs only $k\log_2 n$ bits --- select $k$ features out of $n$ possible
and automatically set their weights to $1/k$.
Next, for a better statistical interpretation
we assume the somewhat idealized case when all features
have equal means $(\mu_P, \mu_N)$ and equal standard deviations
$(\sigma_P, \sigma_N)$ over positive (P) and negative (N) training sets, respectively.


\noindent \textbf{Proposition 2:}
If we assume that the input soft classifiers are independent
and better than random chance, the error rate converges towards $0$
as their number $n$ goes to infinity.

\noindent \textbf{Proof:}
Given a classification threshold $\theta$ for $\mathbf{w}^T\mathbf{f}_i$,
such that $\mu_N < \theta < \mu_P$, then, as $n$ goes to infinity, the probability that a negative sample will have an
average response greater than $\theta$ (a false positive) goes to $0$.
This follows from Chebyshev's inequality.
By a similar argument, the chance of a false negative also goes to $0$ as
$n$ goes to infinity.

\noindent \textbf{Proposition 3:}
The weighted average $\mathbf{w}^T\mathbf{f}_i$
with smallest variance over positives (and negatives)
has equal weights.

\noindent \textbf{Proof:} We consider the case when $\mathbf{f}_i$'s are
from positive
samples, the same being true for the negatives.
Then $\text{Var}(\sum_jw_j\mathbf{f}_i(j)/\sum_jw_j) = \sum w_j^2/(\sum w_j)^2 \sigma^2_P$.
We minimize $\sum w_j^2/(\sum w_j)^2$ by setting its partial
derivatives to zero and get $w_q(\sum w_j)=\sum w_j^2, \forall q$. Then
$w_q=w_j, \forall q,j$.

\begin{figure}[t!]
\begin{center}
\includegraphics[scale = 0.26, angle = 0, viewport = 0 0 900 950, clip]{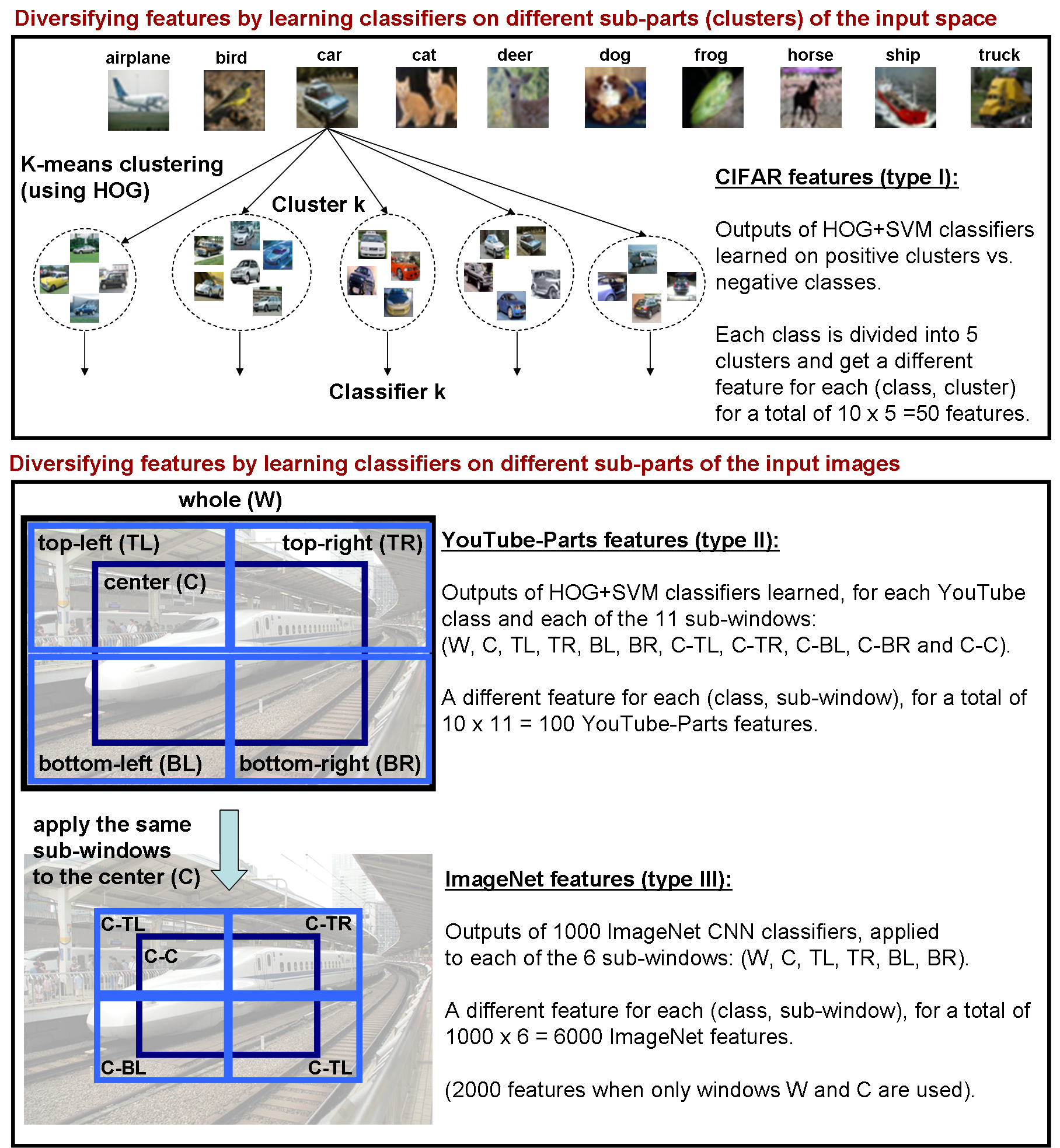}
\caption{We encourage feature diversity by taking classifiers
trained on 3 datasets and by looking at different parts of the input space (Type~I) or different locations within the image (Types~II and~III).
}
\label{fig:diverse_features}
\end{center}
\end{figure}

\section{Experimental Analysis}
\label{sec:experiments}

We evaluate our method's ability to generalize and learn quickly from limited
data, in both the supervised and the unsupervised cases.
We also explore the possibility of transferring and combining
knowledge from different datasets, containing
video or low and medium-resolution images of many potentially
unrelated classes, by working with three different types of features,
as explained shortly. We focus on video classification
and compare to established methods for selection and classification
and report accuracies per frame.
We test on the large-scale YouTube-Objects video
dataset \cite{prest2012learning}, with
difficult sequences from ten categories (aeroplane, bird, boat, car, cat, cow, dog, horse, motorbike, train)
taken \emph{in the wild}. The training set contains about $4200$
video shots, for a total of $436970$ frames, and the test set has $1284$ video shots for a total of over $134119$ frames.
The videos have significant clutter, with objects coming in and out of
foreground focus, undergoing occlusions, extensive changes in scale and viewpoint. This set is difficult because the
\emph{intra}-class variation is large and sudden between video shots. Given the very large number of frames and variety of shots,
their complex appearance and variation in length, presence of background clutter with many distracting objects,
changes in scale, viewpoint and drastic intra-class variation, the task of learning the main category
from only a few frames presents a significant challenge. We used the same training/testing
split as prescribed in \cite{prest2012learning}. In all our tests, we present results averaged over $30$ randomized trials, for each method.
We generate a large pool of over $6000$ different features (see Fig.~\ref{fig:diverse_features}), computed and learned
from three different datasets: CIFAR10 \cite{krizhevsky2009learning}, ImageNet \cite{imagenet_cvpr09}
and a hold-out part of the YouTube-Objects training set:

\paragraph{CIFAR10 features (Type~I):}
This dataset contains
$60000$ 32$\times$32 color images in $10$ classes
(airplane, automobile, bird, cat, deer, dog, frog, horse, ship, truck),
with $6000$ images per class.
There are $50000$ training and $10000$ test images.
We randomly chose $2500$ images per class to create features.
They are HOG+SVM
classifiers trained on data obtained by clustering images from each class
into $5$ groups
using k-means applied to their HOG descriptors.
Each classifier was trained to separate its own cluster
from the others.
We hoped to obtain, for each class, diverse and relatively
independent classifiers that respond to different, naturally clustered, parts of the input space.
Note that CIFAR10 classes coincide only partially ($7$ out of $10$) with the YouTube-Objects classes. Each of the $5\times10=50$ such classifiers becomes a different feature.

\paragraph{YouTube-parts features (Type~II):}
We formed a separate dataset with $25000$ images from video, randomly selected from a subset
of YouTube-Objects training videos, not used in subsequent recognition experiments. Features are outputs of linear SVM classifiers using HOG applied to the different parts of each image.
Each classifier is trained and applied to its own dedicated sub-window
as shown in Fig.~\ref{fig:diverse_features}.

We also applied PCA to the resulted HOG, and obtained descriptors of $46$ dimensions, before passing them to SVM.
For each of the $10$ classes, we have $11$ classifiers, one for
each sub-window, and get a total of $110$ type II features.

\paragraph{ImageNet features (Type~III):}
We considered the soft feature outputs (before soft max) of the pre-trained ImageNet CNN features using Caffe \cite{jia2014caffe},
each of them over six different sub-windows: \emph{whole, center, top-left, top-right, bottom-left, bottom-right},
as presented in Fig.~\ref{fig:diverse_features}. There are $1000$ such outputs, one for each ImageNet category,
for each sub-window, for a total of $6000$ features. In some of our experiments, when specified, we
used only $2000$ ImageNet features, restricted to the \emph{whole} and \emph{center} windows.

\subsection{Results}

\begin{figure}[t!]
\begin{center}
\includegraphics[scale = 0.47, angle = 0, viewport = 0 0 600 840, clip]{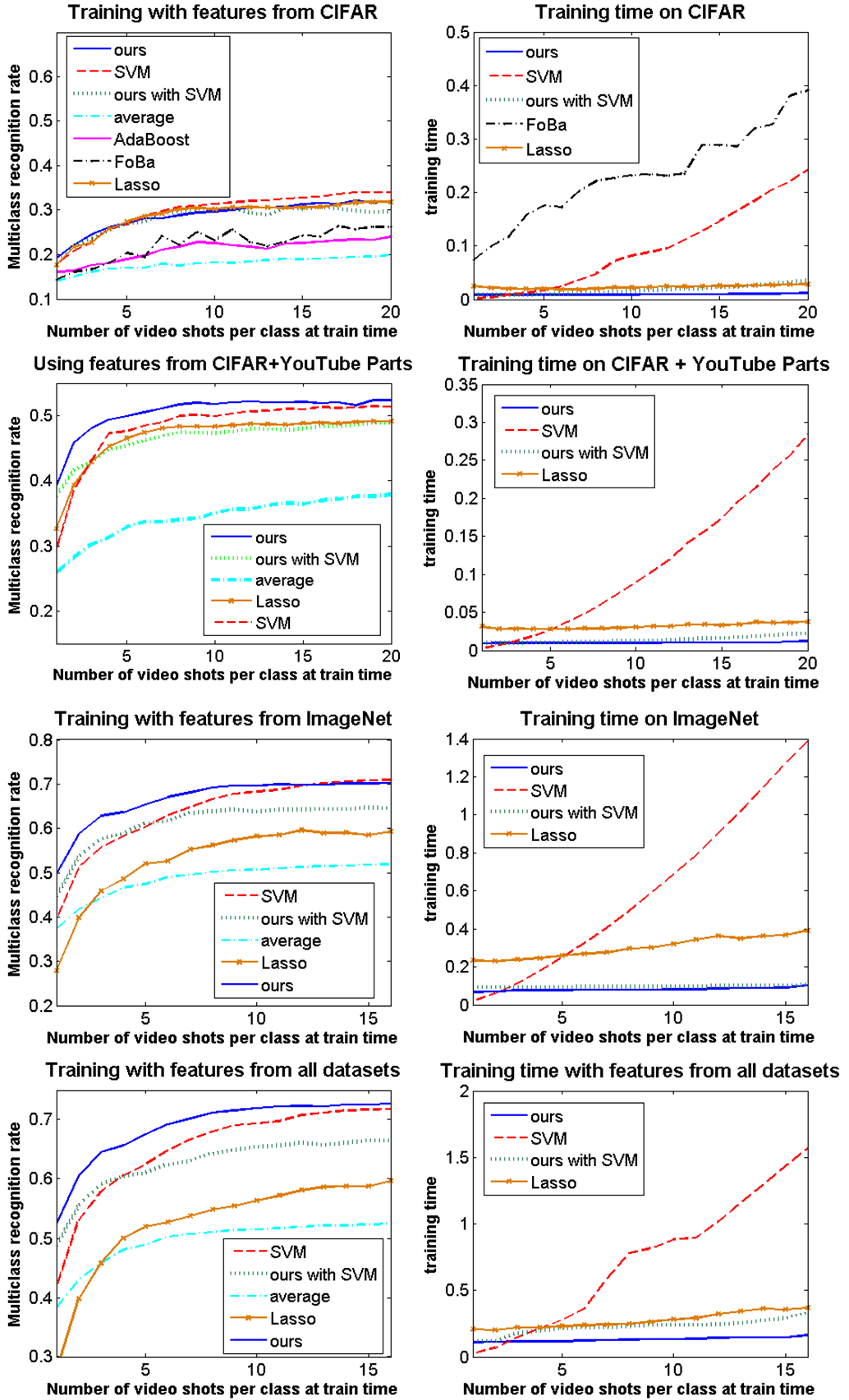}
\caption{Accuracy and training time (in sec.)
on YouTube-Objects, with varying training video shots ($10$ frames per shot and results averaged over $30$ runs).
Input feature pool,
row~1: $50$ type~I features on CIFAR10;
row~2: $110$ type~II features on YouTube-Parts + $50$ CIFAR10;
row~3: $2000$ type~III features in ImageNet;
row~4: $2160$ all features.
Ours outperforms SVM, Lasso, AdaBoost and FoBa.
}
\label{fig:test_and_time}
\end{center}
\end{figure}

\begin{figure}[t!]
\begin{center}
\includegraphics[scale = 0.41, angle = 0, viewport = 0 0 700 480, clip]{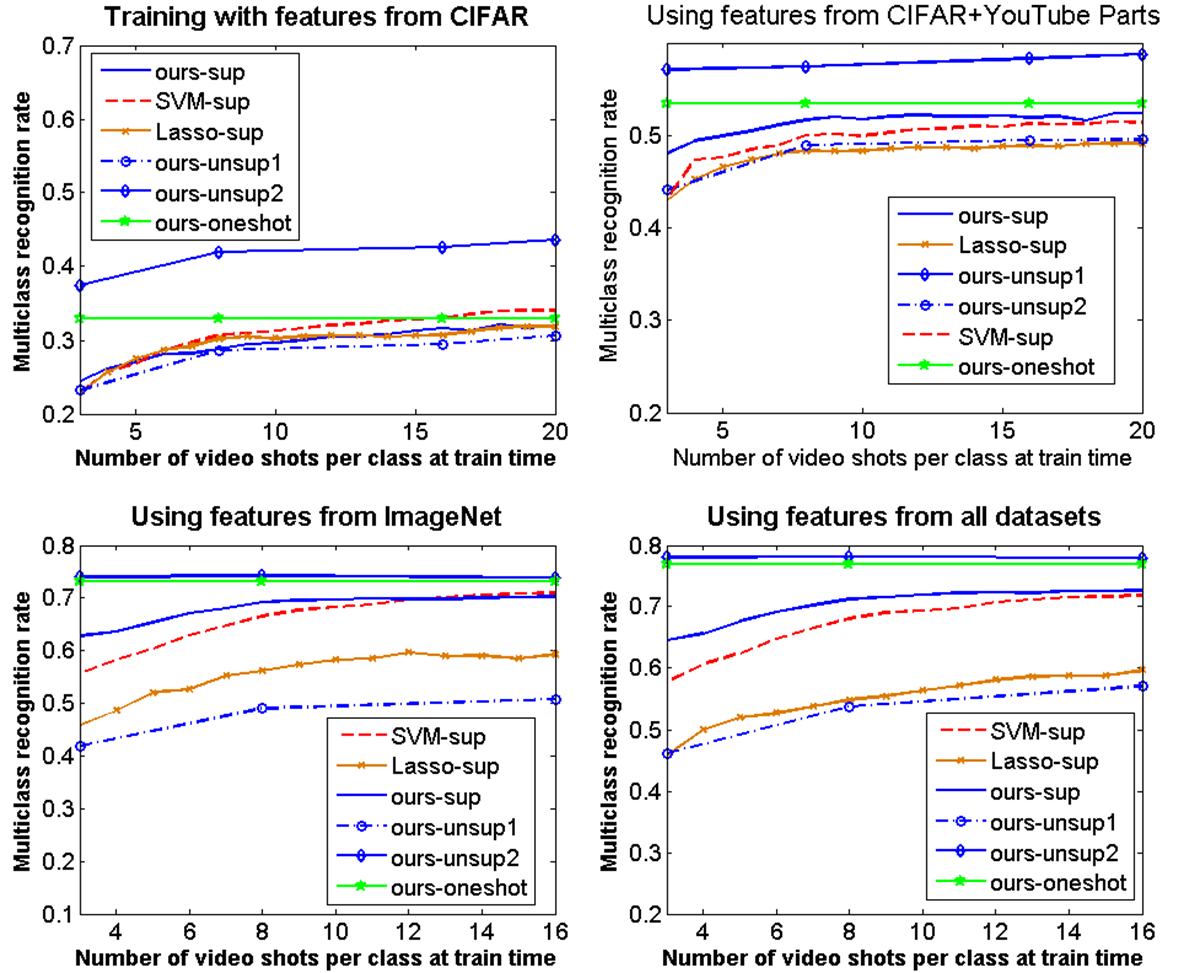}
\caption{Comparison of our almost unsupervised approach
to the supervised case for different methods. In our case, unsup1 uses training data in
$J_u(\mathbf{w})$ only from the shots used
by all supervised methods;
unsup2 also includes frames from testing shots, with unknown test labels; oneshot is unsup2 with a single labeled
image per class used only for feature sign estimation.
This truly demonstrates the ability of our approach to efficiently learn with minimal supervision.
}
\label{fig:semisup}
\end{center}
\end{figure}

We evaluated eight methods: ours,
SVM on all input features, Lasso, Elastic Net (L1+L2 regularization) \cite{zou2005regularization},
AdaBoost on all input features, ours with SVM
(applying SVM only to features selected by our method,
idea related to \cite{nguyen2010optimal,weston2000feature,kira1992feature}),
forward-backward selection (FoBa) \cite{zhang2009adaptive}
and simple averaging of all signed features, with
values in $[0,1]$ and flipped
as discussed before. While most methods work directly with the signed features provided, AdaBoost further transforms each feature into a weak binary classifier by choosing the threshold
that minimizes the expected exponential loss at each iteration (this explains why AdaBoost is much
slower). For SVM we used the LIBSVM \cite{libsvm} implementation version $3.17$, with kernel and parameter $C$ validated separately for each type of experiment.
For the Lasso we used the latest Matlab library and validated
the L1-regularization parameter $\lambda$ for each experiment.
For the Elastic Net we also validated parameter alpha that combines
the L1 and L2 regularizers.
The results (Fig.~\ref{fig:test_and_time}) show
that our method has a constant optimization time (after creating $\mathbf{F}$, and then computing
$\mathbf{F^\top F}$). It is significantly
faster than SVM, AdaBoost (time too large to show in the plot), FoBa and even the latest Matlab's Lasso.
Elastic Net, not shown in the plots to avoid clutter, was consistently slower than Lasso by
at least $35\%$ and, at best, $2\%$ superior in performance to Lasso for certain parameters $alpha$.
As seen, we outperform most other methods, especially in the case of limited labeled training data,
when our selected feature averages generalize well and
are even stronger than in combination with SVM.
In the case of the almost unsupervised learning,
we outperformed all other methods by a very large margin,
up to over $20\%$ (Fig.~\ref{fig:semisup} and Table~\ref{tab:semisup}).
Of particular note
is when only a single labeled image per class was used to estimate the feature signs, with all the other data being unlabeled (Fig.~\ref{fig:semisup}).

\begin{table}
\caption{Improvement
in recognition of unsupervised vs.
the supervised method. Experiments with adding 
unlabeled training data to (1,3,8,16) labeled shots (used for estimating feature signs)
reveals significant improvement over the supervised learning scheme, 
across all trials. The first column presents the one-shot learning case, when
the almost unsupervised method uses a single labeled image per 
class to estimate the feature signs.
Results are averages (in percent) over $30$ random runs.
}
\label{tab:semisup}
\begin{center}
\begin{tabular}{lcccc}
\toprule
Training \# shots   & 1         &  3            & 8             & 16          \\
\midrule
Feature~I           & +15.1   &  +16.9        &  +13.9     &  +14.0    \\
Feature~I+II        & +16.7   &  +10.2        &  +6.2      &  +6.1     \\
Feature~III         & +23.6   &  +11.2        &  +4.9      &  +3.3     \\
Feature~I+II+III    & +24.4   &  +13.4        &  +6.7      &  +5.4     \\
\bottomrule
\end{tabular}
\end{center}
\end{table}

\paragraph{Estimating feature signs from limited data:}

\begin{figure}[h!]
\begin{center}
\includegraphics[scale = 0.3, angle = 0, viewport = 0 0 500 420, clip]{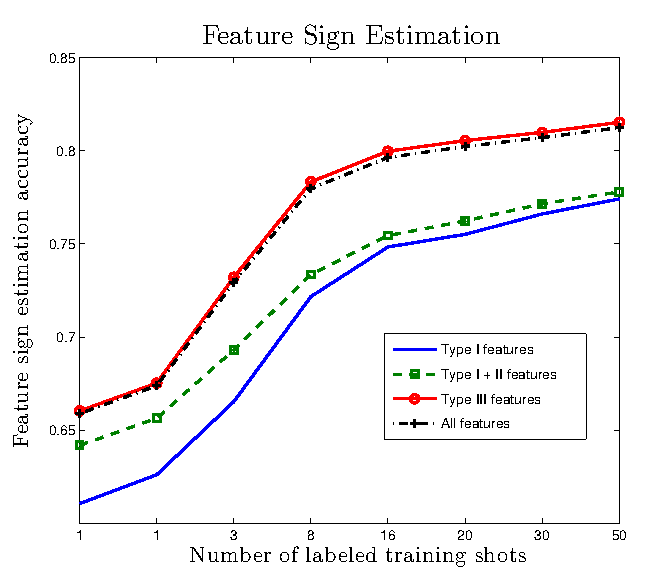}
\caption{For a feature sign estimation accuracy (agreement with sign estimation
from the test set) of roughly $70\%$ our method manages
to significantly outperform the supervised case,
demonstrating its robustness and solid practical value. The first value
corresponds to 1-shot-1-frame case.
}
\label{fig:sign_estimation}
\end{center}
\end{figure}

The performance of
our almost unsupervised learning approach with signed features
depends on the ability to estimate the signs of features.
We evaluate
the accuracy of the estimated signs with respect to
the available labeled data (Fig.~\ref{fig:sign_estimation}).
Our experiments show that feature signs are
often wrongly estimated and thus confirm that our
method is robust to such errors,
with a relatively stable accuracy as the quantity of
labeled samples varies (Fig.~\ref{fig:semisup}).
Note that we have compared the estimated feature signs with the ones estimated from the entire
unlabeled test set of the database
and present estimation \emph{accuracies},
where the signs estimated from the test set were considered the empirical ground truth.
The relatively large sign estimation errors reflect the large relative difference
in quantity between the total amount of test data available and the small number
of samples used for sign estimation. It also indicates our methods ability to learn
effective feature groups in the presence of many others that have been wrongly signed.

\begin{figure}[t!]
\begin{minipage}[b]{1.0\linewidth}
\centering
\includegraphics[width=\textwidth]{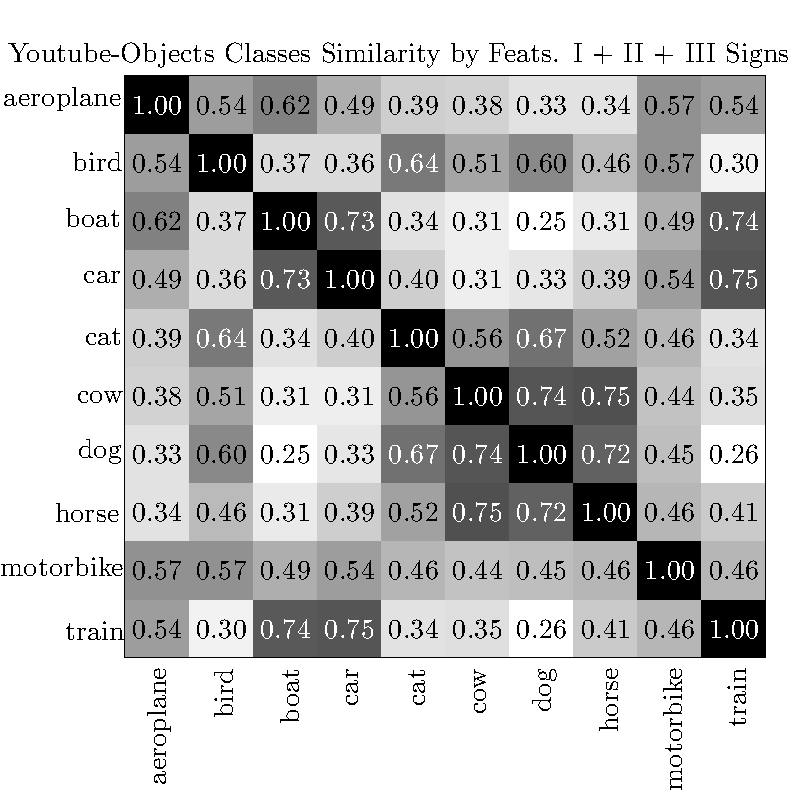}
\end{minipage}
\caption{Youtube-Objects classes similarity based on their estimated signs of features. For each pair of features we present the percent of signs of features that coincide. Note that classes that are more similar in meaning, shape or context
have, on average more signs that coincide. These signs were estimated from all training data.}
\label{fig:class_similarity_by_signs}
\end{figure}

An interesting direction for future work is to explore the possibility of borrowing feature signs from classes that are related
in meaning, shape or context. We have performed some experiments and compared the estimated feature
signs between classes (see Figure \ref{fig:class_similarity_by_signs}). Does the plane share more feature signs with the bird,
or with another man-made class, such as the train? The possibility of sharing or borrowing feature signs from other classes could
pave the way for a more unsupervised type of learning, where we would not need to estimate the signs from labeled data of the specific class.
The results in Figure \ref{fig:class_similarity_by_signs} indicate that, indeed, classes that are closer in meaning
share more signs than classes that mean very different things. For example, the class
\emph{aeroplane} shares most signs with \emph{boat, motorbike, bird, train}, \emph{bird} with \emph{cat, dog, motorbike, aeroplane, cow},
\emph{boat} with \emph{train, car aeroplane}, and \emph{car} with \emph{train, boat, motorbike}. We also have
\emph{cat: dog, bird, cow, horse}, \emph{cow: horse, dog, cat, bird},
\emph{dog: cow, horse, cat, bird}, \emph{horse: cow, dog, cat},
\emph{motorbike: aeroplane, bird, car}, and \emph{train: car, boat, aeroplane}, for the remaining classes. We notice that indeed
classes that are similar in meaning, appearance or context,
such as animals, or man-made categories, share more signs among themselves than classes that are very different.
These experiments indicate the deeper conceptual difference between labeling features and not samples. As our method can be effective
even in case of sign estimation errors, it could relay on some sort of \emph{smart sign guessing} and then
learn from completely unsupervised data - this would reduce the amount of supervision to a minimum, and get closer
to the natural limits of learning in strongly unsupervised environments.


\paragraph{Intuition regarding the selected features:}

\begin{figure*}
\begin{center}
\includegraphics[scale = 0.35, angle = 0, viewport = 0 0 1400 1550, clip]{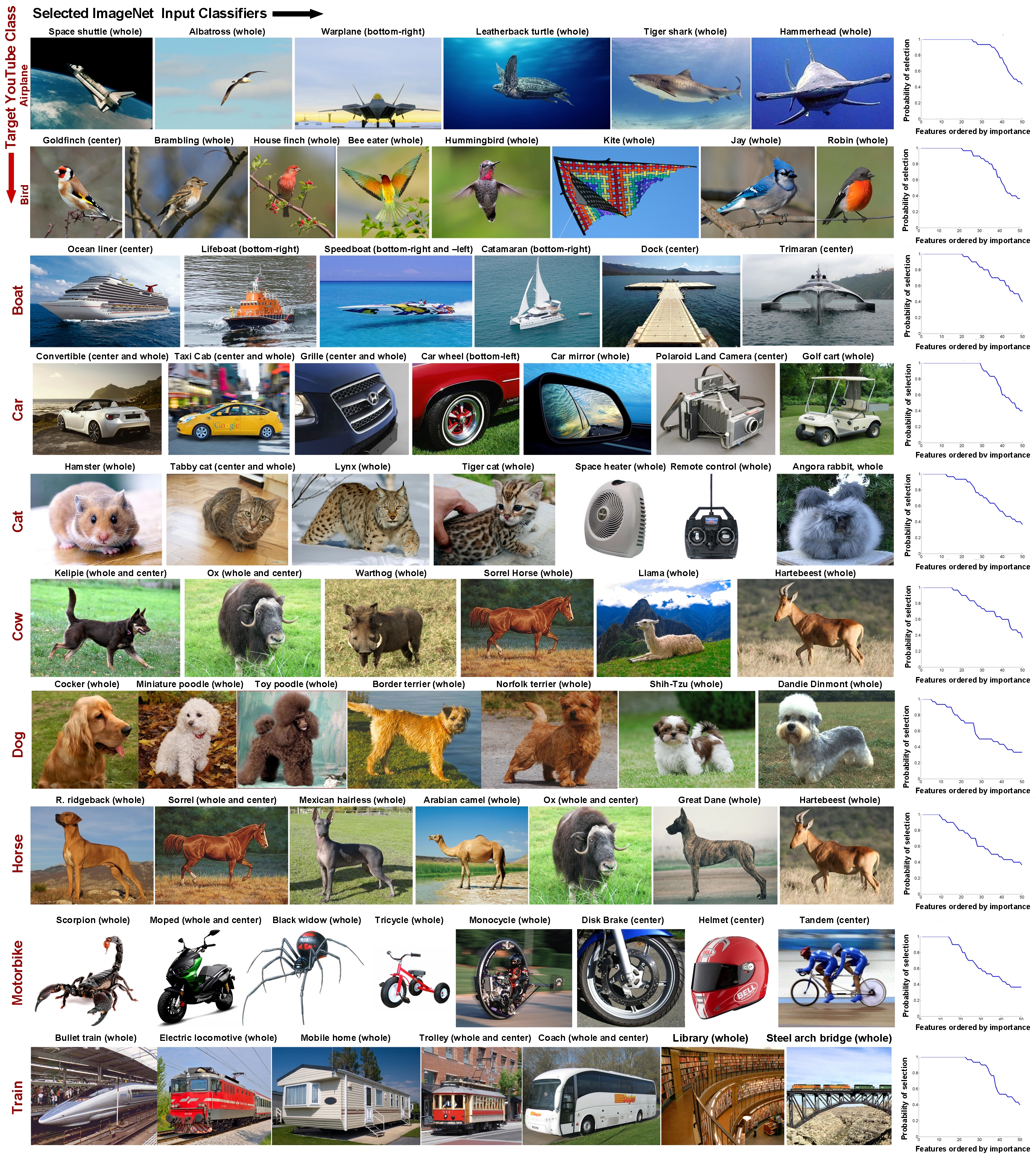}
\caption{Visualization of classifiers selected by our method for each target concept (\emph{not} samples of images retrieved for a given concept). Each row corresponds to a class from the YouTube-Objects dataset and the images in a given row show an image from the most frequently selected ImageNet classifiers (input features) that contribute to that class---specifically the classes that were always selected over 30 independent experiments ($k=50$, 10 frames per shot and 10 random shots for training). The far right graph in each row shows the probability of selecting these 50 features for the given class. Note the stability of the selection process. Also note the connection between the selected ImageNet classifier and the target YouTube object class in terms of appearance, context or geometric part-whole relationships. We find two aspects particularly interesting: 1) the consistent selection of the same classes, even for small random training sets and 2) the fact that semantically unrelated classes contribute to classification, based on their shape and appearance similarity.}
\label{fig:discovered_classes}
\end{center}
\end{figure*}

Another interesting finding (see Fig.~\ref{fig:discovered_classes})
is the consistent selection of diverse
input Type III features that are related to the
target class in surprising ways:
\begin{inparaenum}[1)]
\item similar w.r.t.\
global visual appearance, but not semantic meaning --- banister :: train, tiger shark :: plane, Polaroid camera ::
car, scorpion :: motorbike, remote control :: cat's face, space heater :: cat's head;
\item related in co-occurrence and context, but not in global appearance --- helmet vs.\ motorbike;
\item connected through part-to-whole relationships ---
\{grille, mirror and wheel\} :: car;
or combinations of the above --- dock :: boat, steel bridge :: train,
albatross :: plane.
\end{inparaenum}
The relationships between the target class and the selected features could
also hide combinations of many other factors.
Meaningful relationships could ultimately
join together correlations along many dimensions, from appearance to
geometric, temporal and interaction-like relations.
Since categories share shapes, parts and designs, it is perhaps unsurprising
that classifiers trained on semantically distant classes that are
visually similar can help improve learning and generalization from limited
data.
Another interesting aspect is that the classes
found are not necessarily central to the main category, but often peripheral, acting as guardians that separate
the main class from the rest. This is where feature diversity plays an important role, ensuring both \emph{separation}
from nearby classes as well as robustness to missing values.
This aspect is also related to the idea of borrowing features
from related, previously learned classes.
Thus, in cases where there is insufficient
supervised data for a particular new class, sparse averages of reliable,
old classifiers and features can be an excellent way to combine previous knowledge.
Consider the class \emph{cow} in Fig.~\ref{fig:discovered_classes}.
Although ``cow'' is not present in the $1000$ label set, our method is
able to learn the concept by combining existing classifiers.

\paragraph{Comparison with Linear SVM:}

\begin{figure}[t!]
\begin{center}
\includegraphics[scale = 0.27, angle = 0, viewport = 0 0 900 530, clip]{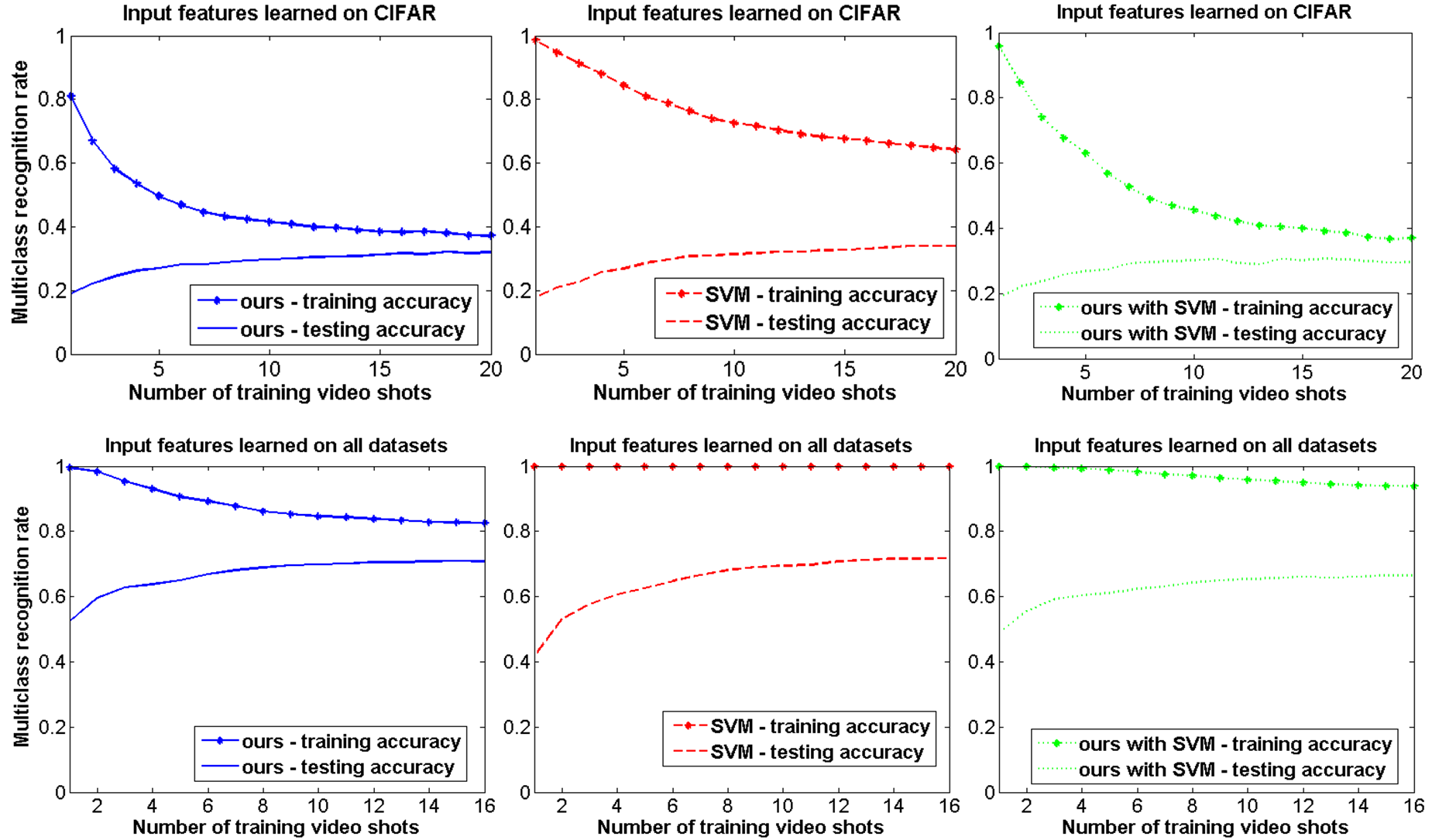}
\caption{Supervised learning: Our method generalizes better (training and test errors are closer) compared to SVM or in combination with SVM
on limited training data.}
\label{fig:train_vs_test}
\end{center}
\end{figure}

\begin{figure}
\begin{center}
\includegraphics[scale = 0.3, angle = 0, viewport = 0 0 700 290, clip]{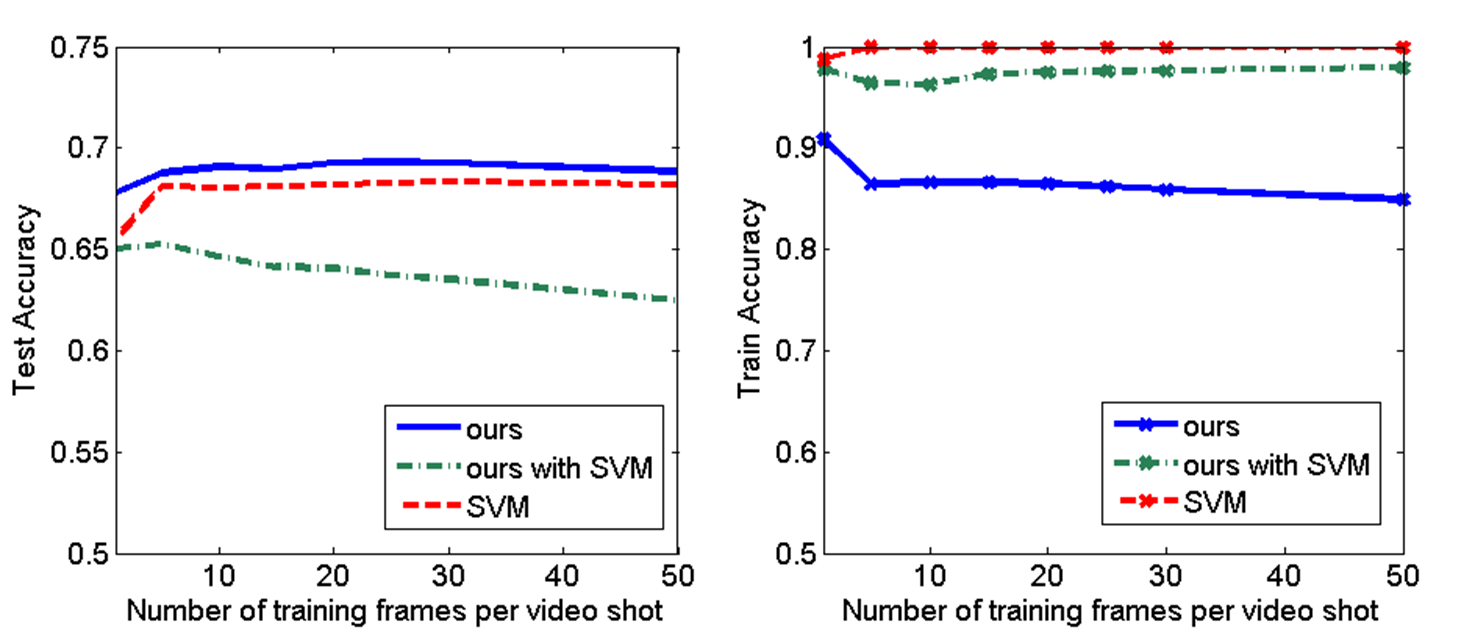}
\caption{Supervised learning: avg. test recognition accuracy over $30$ independent experiments of our method as we vary the number
of training frames uniformly sampled from $8$ random training video shots. Note how well our method generalizes from
just $1$ frame per video shot, for a total of $8$ positive training frames per class.}
\label{fig:accuracy_vs_nFrames}
\end{center}
\end{figure}

In our experiments, the supervised learning 
method generalized significantly better, on average, than SVM or in combination with SVM in cases 
of very limited labeled training data.
We believe that this is due to the power of feature averages, as also indicated by our theoretical results presented earlier. 
Our formulation
is expected to discover features that are independent and strong as a group, not necessarily individually. That is why we prefer to give all selected features equal weight than to put too much faith into a single strong feature, especially in the case of limited training data.
As seen in Figure \ref{fig:train_vs_test}, our supervised approach generalizes better than SVM or in combination with SVM, as reflected by the performance differences between the testing and training cases. Note that we have used a very recent SVM library (libsvm-3.17) with kernel and parameter $C$ validated separately for each type of experiment - in our experiments the linear kernel performed the best.
We can also see that our method often generalizes from just $1$ frame per video shot,
for a total of $8$ positive training frames per class in the experiments in Fig. \ref{fig:accuracy_vs_nFrames}.

\paragraph{Varying the amount of unsupervised data:}

\begin{figure}[h!]
\begin{center}
\includegraphics[scale = 0.35, angle = 0, viewport = 0 0 400 300, clip]{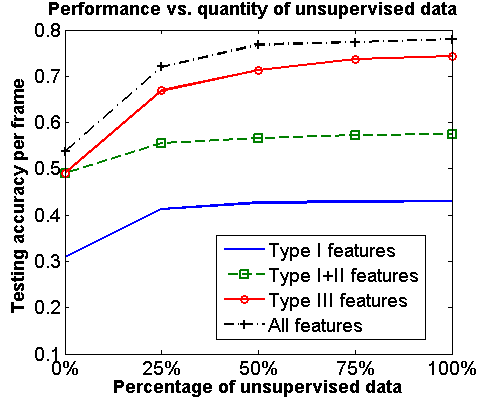}
\caption{Testing accuracy vs amount of unlabeled data used for unsupervised learning.
Note that the performance almost reaches a plateau after a relatively small fraction of unlabeled
frames are added.
}
\label{fig:unsup_amount}
\end{center}
\end{figure}

We evaluated the
influence of varying amounts of unlabeled
data used for the (almost) unsupervised learning method.
We present the following experimental setup:
first, we randomly split the unlabeled test data into two equal sets of frames, $S_A$ and
$S_B$. We have used one set of unlabeled frames $S_A$
for unsupervised learning and the other set $S_B$ for testing.
While keeping the test frames set $S_B$ constant, and $8$ labeled video shots per class for
feature sign estimation from the training set (with
$10$ evenly spaced frames per shot), we varied the amount of  unlabeled
data used for unsupervised learning, by varying the percentage of unlabeled frames used from $S_A$.
We present results over $30$ random runs in Figure~\ref{fig:unsup_amount} and Table~\ref{tab:unsup_amount}.

\begin{table*}[h!]
\caption{Testing accuracy when varying the quantity of unlabeled data used for unsupervised learning. We used $8$ video
shots per class with 10 frames each, for estimating the signs of the features. Results are averages over $30$ random runs.}.

\label{tab:unsup_amount}
\begin{center}
\begin{tabular}{lcccc}
\toprule
Unsupervised data   &   Features~I       &  Features~I+II         & Features~III       & Features I+II+III          \\
\midrule
Train + 0\% test $S_A$   &  30.86\%            &  48.96\%                &  49.03\%        &  53.71\% \\
Train + 25\% test $S_A$   &  41.26\%           &  55.50\%                &  66.90\%         &  72.01\%     \\
Train + 50\% test $S_A$  &  42.72\%           &  56.66\%                &  71.31\%         &  76.78\%     \\
Train + 75\% test $S_A$  &  42.88\%          &  57.24\%                &  73.65\%          &  77.39\%     \\
Train + 100\% test $S_A$ &  43.00\%          &  57.44\%                &  74.30\%          &  78.05\%     \\
\bottomrule
\end{tabular}
\end{center}
\end{table*}

\section{Discussion and Conclusions}

We present a fast feature selection and
learning method with minimal supervision,
and we apply it to video classification.
It has
strong theoretical properties and
excellent generalization and accuracy in practice.
The crux of our approach is its ability to learn from large quantities of
unlabeled data once the feature signs are determined, while being very robust
to feature sign estimation errors.
A key difference between our features signs and the weak features used by boosting
approaches such as AdaBoost, is that in our case the sign estimation requires minimal labeling
and that \emph{the sign} is the only bit of supervision needed.
Adaboost requires large amounts of training
data to carefully select and weigh new features.
This aspect reveals a key insight:
being able to approximately
label \emph{the features} and \emph{not the data}, is sufficient for learning.  
With a formulation that permits very fast optimization
and effective learning
from large heterogeneous feature pools,
our approach provides a useful tool for many other recognition tasks,
and it is suited for real-time, dynamic environments.
Thus it could
open doors for new and exciting research in machine
learning, with both practical and theoretical impact.\\
\textbf{Acknowledgements:}
This work was supported in part by CNCS-UEFICSDI,
under project PNII PCE-2012-4-0581.


\bibliography{complete}
\bibliographystyle{aaai}
\end{document}